\documentclass[letterpaper,10pt,conference]{ieeeconf}

\IEEEoverridecommandlockouts
\usepackage{amsmath}
\usepackage{amssymb}
\usepackage{graphicx}
\usepackage{multirow}
\usepackage{booktabs}
\usepackage{tabularx}

\title{\LARGE \bf
Self-Supervised Consistency Enhanced Disentangled Learning for Neural Decoding Generalization in Brain-Machine Interfaces
}

\author{Jiyu Wei$^{1,2}$, Di Hong$^{1,2}$, Zhanjie Zhang$^{1}$, Dazhong Rong$^{*,1,2}$, Qinming He$^{1,2}$, Yueming Wang$^{*,1,2}$
\\$^{1}$College of Computer Science and Technology, Zhejiang University, Hangzhou, China
\\$^{2}$Nanhu Brain-Computer Interface Institute, Hangzhou, China
\\{\tt\{weijiyu,hongd,cszzj,rdz98,hqm,ymingwang\}@zju.edu.cn}
\thanks{*Corresponding authors: Dazhong Rong and Yueming Wang.}
}

\begin{document}

\maketitle
\thispagestyle{empty}
\pagestyle{empty}

\begin{abstract}

Brain–Machine Interfaces (BMIs) provide a direct communication pathway between the brain and external devices, enabling humans to control assistive and robotic technologies, with potential applications in rehabilitation, human motor augmentation, and human-centered robotics. However, due to neural drift, the performance of BMIs decreases over time, posing challenges for long-term viability, particularly for invasive BMIs (iBMIs). Existing solutions suffer from two main drawbacks: (i) difficulty in learning robust neural representations, and (ii) neglecting that neural drift varies across motor parameters (\textit{e.g.,} velocity, direction, and speed). To overcome these limitations, we propose Self-Supervised Consistency enhanced Disentangled Learning (SSCDL), a neural decoding generalization framework built on two key innovations. We first design a backbone model named Consistency enhanced Neural Decoder (CND), using a novel teacher-student consistency constraint with simulated neural signal perturbations to learn robust representations invariant to neural drift. Then, we employ three dedicated CNDs under Complementary-Disentangled Generalization (CDG) mechanism, which disentangles motor signals into velocity, direction and speed with the inspiration of neural preference theory. This disentangled learning enables SSCDL to capture invariant neural representations from diverse neural preference perspectives, significantly enhancing cross-day generalization. Extensive experimental results show that SSCDL delivers state-of-the-art decoding performance, exhibiting high robustness and cross-day stability. These capabilities underscore its strong potential for long-term interaction in human-centric robotic and fine-grained assistive applications.

\end{abstract}

\section{Introduction}
Brain–Machine Interfaces (BMIs) translate neural activity into actionable commands for external devices, thereby enabling direct interaction with assistive and robotic technologies~\cite{ICRA2016,ICRA_Rehi,ICRA_EEG_motor}. Among them, invasive motor BMIs (iBMIs) leverage high-quality neural recordings from motor cortical areas, providing not only a powerful assistive tool but also a unique experimental platform for investigating the neural mechanisms underlying motor control~\cite{bci1,bci2,bci3,restore1}. Owing to these advantages, iBMIs have demonstrated significant potential in neurorehabilitation, human motor augmentation, and human-centered robotics.

\begin{figure}[t]
    \centering
    \includegraphics[width=0.99\linewidth]{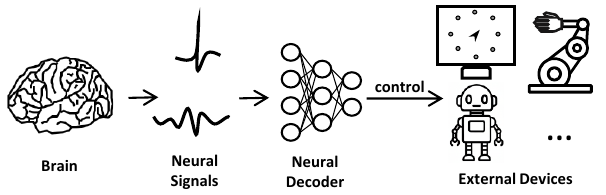}
    \caption{Schematic of a Brain–Machine Interfaces (BMIs): neural activity is recorded from the brain, processed by a decoder, and converted into control signals for external devices such as robots, computers, and assistive systems.}
    \label{fig:bmi}
\end{figure}

As shown in Fig.~\ref{fig:bmi}, a central component of iBMIs is neural decoder, which maps recorded brain activity into control commands through sophisticated decoding algorithms~\cite{PD,speed2,KRSRL,DyEnsemble,EvoDyEnsemble,gao2025cross}. Although a variety of decoding methods have achieved reliable short-term performance, their accuracy typically degrades over days due to neural drift~\cite{long,UAN}. This variability stems from factors such as electrode drift, neuronal turnover, and synaptic plasticity~\cite{Brainplasticity,Brainplasticity2}, substantially hindering the widespread adoption and practical application of iBMIs.

Current solutions to address neural drift can be broadly categorized into recalibration-based and generalization-based methods. Recalibration-based methods~\cite{daily,wen2023rapid,adan,UAN,WDGRL,SeSA} maintain long-term decoding performance by retraining decoders with newly collected labeled or unlabeled data. However, this reliance on fresh data collection is often impractical in real-world applications, where acquiring such data is both costly and time-consuming. In contrast, generalization-based methods improve cross-day robustness without requiring decoder retraining, which is also the focus of this work. Representative generalization-based methods include: KRSRL~\cite{KRSRL} and EvoEnsemble~\cite{EvoDyEnsemble} model short-term neural signal changes by siamese learning and ensemble strategies; SABLE~\cite{SABLE} and CAPTIVATE~\cite{CAPTIVATE} employ adversarial feature alignment and neuron-based latent synchronization; MRNN~\cite{MRNN} models long-term nonlinear neural-kinematic mappings; and LFDA~\cite{LFDA} leverages data augmentation with contrastive learning to simulate temporal variations. Despite these advances, learning long-term stable neural representations from only single-day data remains highly challenging. 

\begin{figure}[t]
    \centering
    \includegraphics[width=0.99\linewidth]{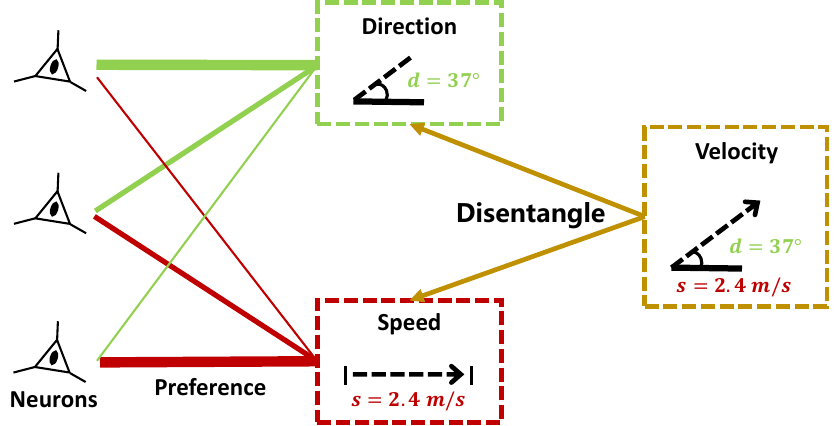}
    \caption{Neural tuning preference: individual neurons exhibit distinct selectivity to different motor parameters disentangled from velocity (\textit{e.g.,} speed and direction), implying that neural drift may vary across these parameters. Line thickness indicates the strength of preference.}
    \label{fig:preference}
\end{figure}

One key limitation of existing generalization-based methods is that they treat velocity as a unified target variable, overlooking the fact that neural drift may manifest differently across motor parameters. Specifically, since individual neurons exhibit distinct tuning preferences for direction and speed~\cite{PD, ps}, the drift affecting neural representations may not be uniform across these dimensions (as shown in Fig.~\ref{fig:preference}). Recent studies~\cite{SeSA} further suggest that aligning velocity features as a whole can lead to suboptimal alignment in its constituent components, such as speed and direction. Consequently, enforcing uniform generalization across the entire velocity space may compromise the model generalization ability on specific motor parameters.

In this paper, we argue that disentangling neural decoding into velocity and its constituent components (direction and speed) provides a more effective strategy for capturing invariant neural representations and mitigating neural drift, thereby improving cross-day generalization. Rather than treating velocity as a unified target, this parameter-aware decomposition explicitly accounts for the heterogeneous tuning preferences of neurons. Our hypothesis is conceptually aligned with recent advances in disentangled representation learning~\cite{tran2017disentangled,zhang2022towards}, which have demonstrated strong potential for extracting robust and transferable features under distribution shifts.

To address the aforementioned limitations, we propose Self-Supervised Consistency enhanced Disentangled Learning (SSCDL), a robust neural decoding framework for long-term brain-machine interface in human-centered robotic systems. Building on recent progress in self-supervised learning (SSL)~\cite{MAE,SSL_survey,rong2025improving}, we first develop a Consistency enhanced Neural Decoder (CND) based on a teacher-student paradigm as the backbone model. CND enforces a representation-level consistency constraint, encouraging the teacher and student networks to produce aligned outputs under simulated neural signal perturbations. This design improves the robustness of learned representations against neural variability across days, even when training is limited to single-day recordings. Motivated by the well-established principle that individual neurons exhibit distinct preferences for motor parameters~\cite{ps,PD,SeSA,CDNG}, we further introduce Complementary-Disentangled Generalization (CDG) and employ our CNDs under it. Specifically, CDG decomposes conventional neural decoding into three interconnected components---overall velocity, direction, and speed---and assigns a dedicated SSD to each component. The resulting predictions are integrated through a complementary ensemble mechanism, enabling the model to capture invariant representations across multiple motor dimensions. This structured design facilitates more precise and stable control, which is particularly critical for long-term, fine-grained interaction with assistive and robotic devices. Together, CND and CDG constitute a scalable and flexible generalization framework that mitigates cross-day performance degradation while remaining readily extensible to other neural decoding scenarios. Our main contributions are summarized as follows:

\begin{itemize}
\item We propose Self-Supervised Consistency enhanced Disentangled Learning (SSCDL), a novel neural decoding generalization framework that mitigates cross-day decoding performance degradation, enabling long-term stable BMI control in robotic systems.
\item We develop a backbone model, Consistency enhanced Neural Decoder (CND), enforcing representation-level consistency under neural signal perturbations to learn invariant features from single-day data. We apply CND to Complementary-Disentangled Generalization (CDG), which decomposes neural decoding into velocity, direction, and speed components, addressing motor-parameter-dependent neural drift by performing motor-parameter-aware consistency learning.
\item Extensive experiments on multiple iBMI datasets show that SSCDL achieves superior cross-day decoding performance and improved fine-grained prediction accuracy in direction and speed, highlighting its effectiveness for long-term stable BMI control.
\end{itemize}

\section{Related Work}
Due to neural data drift, BMIs face a significant challenge in maintaining decoding accuracy across days. Existing solutions can be broadly categorized into recalibration-based methods and generalization-based methods.

\subsection{Recalibration-based Methods}
Traditional approaches typically rely on periodic decoder retraining using newly collected labeled data~\cite{daily}. However, the collection of labeled data is often costly, severely limiting the practicality of BMIs. To address this, more advanced methods adopt Unsupervised Domain Adaptation (UDA), treating $\mathtt{day\_0}$ data as the source domain and subsequent $\mathtt{day\_k}$ data as the target domain. Among them, ADAN~\cite{adan} and UAN~\cite{UAN} adapt the target domain to the source domain by adversarial learning; DA-DCF~\cite{DA-DCF} and WDGRL~\cite{WDGRL} align the source and target domains from the perspective of feature alignment; SeSA~\cite{SeSA} introduce semi-supervised domain adaptation by integrating semi-supervised learning with UDA to conditionally align speed features between the source and target domains. Despite their effectiveness, these methods still require access to unlabeled or partially labeled data from each new session. This dependency poses practical challenges in real-world BMI deployments.

\begin{figure*}[t]
\centerline{\includegraphics[width=0.99\linewidth]{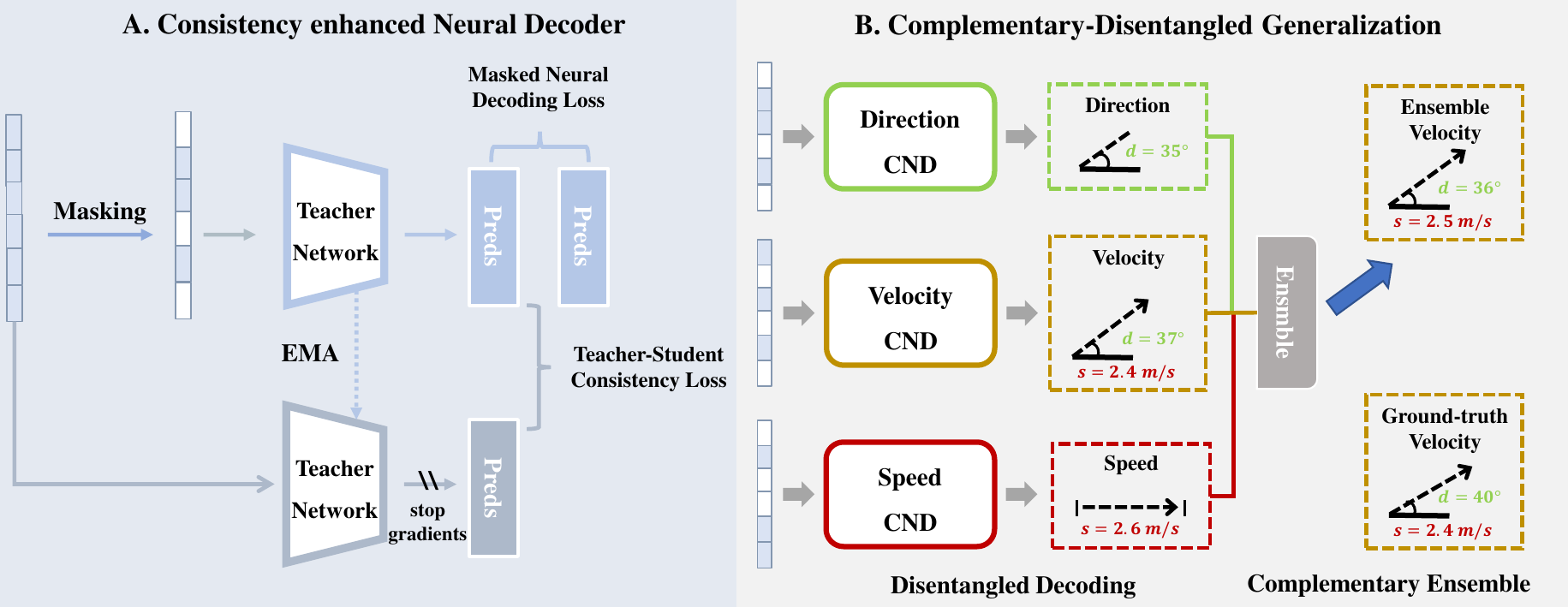}}
\caption{The overall framework of Self-Supervised Consistency enhanced Disentangled Learning.} 
\label{fig:SSCDL}
\end{figure*}

\subsection{Generalization-based Methods}
Generalization-based methods aim to maintain cross-day performance without further model updates. To achieve the goal, KRSRL~\cite{KRSRL} employs a siamese learning framework; DyEnsemble and EvoEnsemble~\cite{EvoDyEnsemble} use ensemble strategies to adapt to short-term variations; CAPTIVATE~\cite{CAPTIVATE} incorporates a neuron locator for feature synchronization; SABLE~\cite{SABLE} leverages adversarial alignment to learn invariant latent dynamics; LFDA~\cite{LFDA} strengthens generalization by data augmentation and contrastive learning; MRNN~\cite{MRNN} models multiple nonlinear mappings between neural signals and kinematics, leveraging historical data to capture long-term variability. However, a significant gap still remains: these methods typically ignore the heterogeneity of neural drift, which is often coupled with motor parameters due to the diverse tuning properties of individual neurons.

\section{Methods}
In this section, we first formally define the problem of cross-day neural decoding. Then, inspired by recent advances in Self-Supervised Learning (SSL), we propose a strong neural decoding backbone termed Consistency enhanced Neural Decoder (CND), which leverages data augmentation to simulate cross-day neural drift and learn robust neural representations. Building upon neural preference theory and the success of Disentangled Representation Learning (DRL) in domain generalization, we further introduce Complementary-Disentangled Generalization (CDG). Lastly, by integrating CND and CDG, we present the overall framework of our Self-Supervised Consistency enhanced Disentangled Learning (SSCDL), as shown in Fig.~\ref{fig:SSCDL}.

\subsection{Problem Definition}
This paper addresses the challenge of maintaining stable BMI decoding performance across days using only single-session training data. We formulate this problem as a single-domain generalization task. Specifically, data collected on $\mathtt{day\_0}$ serve as the source domain $\mathcal{D}_s = \{(x_i, y_i)\}_{i=1}^{N_s}$, while data from a subsequent day $\mathtt{day\_k}$ ($k>0$) constitute an unseen target domain $\mathcal{D}_t$. Here, $x_i$ denotes the recorded neural signal and $y_i$ represents the corresponding kinematic variable. Our goal is to train a decoder $F$ using only $\mathcal{D}_s$ such that it generalizes effectively to $\mathcal{D}_t$, thereby minimizing performance degradation caused by neural drift.

\subsection{Consistency enhanced Neural Decoder}
To learn neural representations robust to cross-day variability, we propose Consistency enhanced Neural Decoder (CND), a strong backbone model built upon SSL. CND adopts a teacher–student architecture, where the student network is optimized using both a supervised decoding loss and a self-supervised consistency loss, while the teacher network provides a stable reference for representation learning.

Let $f_{\theta}$ and $f_{\hat{\theta}}$ denote the student and teacher networks, respectively. At initialization, the parameters of the two networks are identical and randomly initialized. The training of the student network involves the following two losses:
\begin{enumerate}
    \item \textbf{Masked Neural Decoding Loss.} To simulate realistic perturbations such as electrode instability and channel dropout, we introduce a data augmentation strategy termed Neural Signal Masking (NSM). During training, we randomly mask $c$ channels of neural signals $x_i$ in $\mathcal{D}_s$ to produce an augmented input $\hat{x}_i$. The loss for predicting the corresponding kinematic variable $y_i$ from the masked neural signal $\hat{x}_i$ is defined as:
    \begin{equation}
    \mathcal{L}_{\text{MND}}(Z)=\frac{1}{N_s}\sum_{(x_i,y_i)\in\mathcal{D}_s}Z\big(f_{\theta}(\hat{x}_i), y_i \big),
    \end{equation}
    where $Z$ denotes a function that measures the prediction error.
    \item \textbf{Teacher–Student Consistency Loss.} In addition to the decoding objective, we enforce a teacher–student consistency constraint to further enhance representation robustness. The consistency loss is defined as:
    \begin{equation}
    \mathcal{L}_{\text{TSC}}(Z)=\frac{1}{N_s}\sum_{(x_i,y_i)\in\mathcal{D}_s}Z\big(f_{\theta}(\hat{x}_i), f_{\hat{\theta}}(x_i)\big).
    \end{equation}
    The key idea is that the student network, when given perturbed inputs, should produce predictions consistent with those of the teacher network operating on the clean neural signal. This constraint encourages the student network to focus on stable neural structures rather than overfitting to specific neurons. 
\end{enumerate}
The student network is optimized using the combined loss:
\begin{equation}
\mathcal{L}_{\text{CND}}(Z)=\mathcal{L}_{\text{MND}}(Z)+\lambda\cdot\mathcal{L}_{\text{TSC}}(Z),
\end{equation}
where $\lambda$ balances the two loss terms.

As for the teacher network, its parameters are not directly optimized by gradient descent. Instead, its parameters are updated as an Exponential Moving Average (EMA) of the student parameters. Let $k$ denote the training iteration, and $\tilde{k}$ denote a warm-up threshold before which the teacher parameters directly follow the student parameters. The update rule is defined as follows:
\begin{equation}
\hat{\theta}^k=\begin{cases}
\theta^k, & \text{if\quad} k\leq\tilde{k}, \\
\frac{k-\tilde{k}}{k-\tilde{k}+1}\cdot\hat{\theta}^{k-1}+\frac{1}{k-\tilde{k}+1}\cdot\theta^k, & \text{otherwise.}
\end{cases}
\end{equation}
This EMA update produces a temporally smoothed teacher model that provides a more stable reference for consistency learning, which is particularly beneficial for noisy neural recordings.

\subsection{Complementary-Disentangled Generalization}
Neuroscience studies indicate that different neural populations encode distinct motor parameters, such as preferred direction and preferred speed~\cite{PD,ps}. This suggests that neural drift may affect these motor parameters differently. Motivated by this observation, we propose Complementary-Disentangled Generalization (CDG), which improves cross-day generalization ability by explicitly modeling different motor parameters rather than treating velocity as a single unified prediction target. CDG involves two steps as follows:
\begin{enumerate}
    \item \textbf{Disentangled Learning.} In CDG, velocity decoding is decomposed into three complementary components: overall velocity, direction, and speed. Instead of directly learning velocity as a unified output, we train separate models for each of the three components. This disentanglement allows the decoder to capture parameter-specific neural representations that may generalize differently across days. To measure the prediction discrepancy for different motor parameters, we define parameter-specific difference functions:
    \begin{align}
    Z_{\text{direction}}(\hat{y}_i,y_i) &= \|y_i\|\cdot(1-\frac{\hat{y}_i\cdot y_i}{\|\hat{y}_i\|\cdot\|y_i\|}),\\
    Z_{\text{speed}}(\hat{y}_i,y_i) &= (\|\hat{y}_i\|-\|y_i\|)^2,\\
    Z_{\text{velocity}}(\hat{y}_i,y_i) &= \|\hat{y}_i-y_i\|^2.
    \end{align}
    These parameter-specific functions allow each model to focus on learning the neural representations most relevant to its corresponding motor parameter.
    \item \textbf{Complementary Ensemble.} After obtaining predictions for direction, speed, and velocity, we combine them using a complementary ensemble strategy to produce the final decoding result. Let $\hat{y}_i^{\text{direction}}$, $\hat{y}_i^{\text{speed}}$, and $\hat{y}_i^{\text{velocity}}$ denote outputs of the three models, respectively. The final velocity prediction is computed as:
    \begin{equation}
    \tilde{y}_i=\frac{1}{2}\Big(\|\hat{y}_i^{\text{speed}}\|\cdot\frac{\hat{y}_i^{\text{direction}}}{\|\hat{y}_i^{\text{direction}}\|}+\hat{y}_i^{\text{velocity}}\Big).
    \end{equation}
    The first term reconstructs velocity from the disentangled direction and speed predictions, while the second term corresponds to the holistic velocity prediction. By averaging these two complementary estimates, the ensemble leverages both parameter-specific specialization and holistic decoding, resulting in more accurate and robust predictions.
\end{enumerate}

\subsection{Overall Framework}
To realize the introduced CDG mechanism, we employ Consistency enhanced Neural Decoder (CND) as the backbone model for each disentangled branch. Specifically, we construct three independent CND models to decode direction, speed, and overall velocity, respectively. Let $f_{\theta_d}$, $f_{\theta_s}$ and $f_{\theta_v}$ denote the corresponding student networks. Let $f_{\hat{\theta}_d}$, $f_{\hat{\theta}_s}$ and $f_{\hat{\theta}_v}$ denote their teacher networks. Each branch is trained following the CND learning strategy described previously, where the loss functions of the three student networks are $\mathcal{L}_{\text{CND}}(Z_{\text{direction}})$, $\mathcal{L}_{\text{CND}}(Z_{\text{speed}})$ and $\mathcal{L}_{\text{CND}}(Z_{\text{velocity}})$, respectively. For a given neural signal $x_i$, the teacher networks produce predictions:
\begin{align}
    \hat{y}_i^{\text{direction}} &= f_{\hat{\theta}_d}(x_i), \\
    \hat{y}_i^{\text{speed}} &= f_{\hat{\theta}_s}(x_i), \\
    \hat{y}_i^{\text{velocity}} &= f_{\hat{\theta}_v}(x_i).
\end{align}

Since the three branches are fully decoupled, they can be trained in parallel without introducing additional computational overhead. This modular design enables the framework to learn parameter-specific neural representations while maintaining training efficiency. After training, the teacher networks are used for inference. Their predictions are integrated using the complementary ensemble strategy defined before, yielding the final decoding output $\tilde{y}_i$.

\section{Experiments}
In this section, we first describe the experimental setup and then present a series of experiments designed to answer the following research questions (RQs):
\begin{itemize}
\item \textbf{RQ1:} How does our SSCDL compare with other state-of-the-art approaches?
\item \textbf{RQ2:} How do individual modules in SSCDL contribute to the overall performance?
\item \textbf{RQ3:} How robust is SSCDL under different experimental conditions?
\end{itemize}

\subsection{Experimental Settings}
\subsubsection{Datasets}
We evaluate our method on three non-human primate (NHP) datasets from UAN~\cite{UAN}, namely Chewie, Mihili, and Jango. These datasets contain neural recordings collected during various upper-limb motor tasks, including center-out reaching (CO), random-target reaching (RT), and isometric wrist movements (ISO). Neural signals were recorded from the primary motor cortex (M1) using a 96-channel Utah electrode array implanted in the hand or arm representation area of M1. The signals were digitized and bandpass filtered between 250 Hz and 5000 Hz. Multiunit threshold crossings from the go cue time to the trial end time were extracted. To obtain smoothed firing rates, spike counts were binned using 50 ms non-overlapping windows and convolved with a Gaussian kernel with a standard deviation of 100 ms. The resulting firing rates were then z-score normalized. For cross-day generalization experiments, trials from the first recording day are used as the source domain, while trials from subsequent days are treated as target domains. This setting allows us to evaluate the robustness of neural decoders under neural drift across days.

\subsubsection{Implementation Details}
The student and teacher networks of the proposed CND share the same architecture, implemented as a three-layer multilayer perceptron (MLP) with ReLU activations and dropout layers ($p=0.25$). The MLP serves as the feature extractor, followed by a linear regression head that outputs the predicted kinematic variables. During training, the batch size is set to $128$. We use the Adam optimizer with a learning rate of $5\times10^{-3}$, $\textit{betas}=(0.9,0.999)$, and $\textit{weight\_decay}=5\times10^{-4}$. For neural signal masking, the number of masked channels $c$ is set to $32$ for the Chewie dataset and $16$ for the Mihili and Jango datasets. The balance weight $\lambda$ is set to $1$. All experiments are repeated with five random seeds, and the reported results are averaged across these runs to ensure reliability and reduce the effect of stochastic variations.

\subsubsection{Metrics}
We evaluate decoding performance using two widely adopted metrics: Correlation Coefficient (CC) and Coefficient of Determination (R$^2$). For clarity, both metrics are reported in percentage form. In addition, we introduce two auxiliary metrics to evaluate the accuracy of the disentangled motor parameters. For speed prediction, we use Speed Error (SE), defined as the mean squared difference between the predicted and ground-truth speeds. For direction prediction, we use Angle Error (AE), defined as the mean angular difference (in degrees) between the predicted and ground-truth directions. The angular difference is computed using the minimal circular distance to account for the periodic nature of angles (e.g., $0^\circ$ and $360^\circ$ represent the same direction). These metrics provide further insight into the model ability to accurately decode different motor components.

\subsection{Comparison with Other Methods (RQ1)}
To evaluate the generalization capability of SSCDL, we conduct cross-day experiments using four recording days from each dataset. The earliest day (e.g., C0, J0, M0) is used as the source domain, while the subsequent days (C1--C3, J1--J3, M1--M3) serve as target domains spanning short (about one day), medium (about one month), and long (about three months) temporal intervals. Detailed descriptions of these sessions are provided in Table~\ref{tab:dataset}. For comparison, we include a MLP baseline (denoted as Vanilla in Table~\ref{tab:exp-main}), several state-of-the-art generalization-based methods (KRSRL~\cite{KRSRL}, MRNN~\cite{MRNN}, and LFDA~\cite{LFDA}), and a recalibration-based method (WDGRL~\cite{WDGRL}). Note that the WDGRL results are obtained with retraining using target-domain data.

\begin{table}[t]
\caption{Descriptions of Used Recording Sessions}\label{tab:dataset}
\centering
\begin{tabular}{cccc}
\toprule
Dataset & Task & Session ID & Recording Date \\
\midrule
\multirow{4}{*}{Chewie} & \multirow{4}{*}{CO} & C0 & 2016-09-27 \\
& & C1 & 2016-09-28 \\
& & C2 & 2016-10-12 \\
& & C3 & 2016-11-03 \\
\midrule
\multirow{4}{*}{Jango} & \multirow{4}{*}{ISO} & J0 & 2015-07-30 \\
& & J1 & 2015-07-31 \\
& & J2 & 2015-08-20 \\
& & J3 & 2015-10-29 \\
\midrule
\multirow{4}{*}{Mihili} & \multirow{4}{*}{RT} & M0 & 2013-12-07 \\
& & M1 & 2013-12-08 \\
& & M2 & 2014-01-15 \\
& & M3 & 2014-02-24 \\
\bottomrule
\end{tabular}
\end{table}

\subsubsection{Quantitative Results}
Table~\ref{tab:exp-main} reports the cross-day decoding performance of different methods across three datasets and three time spans. Overall, SSCDL consistently achieves the best or highly competitive results. In particular, it attains the highest CC and R$^2$ values in most settings, indicating strong robustness to neural drift. Notably, SSCDL even surpasses WDGRL, which utilizes unlabeled target-domain data for recalibration. This result highlights the effectiveness of our single-domain generalization framework.

\begin{table*}[t]
\centering
\caption{Comparison of Cross-day Generalization Performance}\label{tab:exp-main}
\begin{tabularx}{0.99\linewidth}{ccX<{\centering}X<{\centering}X<{\centering}X<{\centering}X<{\centering}X<{\centering}X<{\centering}X<{\centering}X<{\centering}X<{\centering}X<{\centering}X<{\centering}}
\toprule
\multirow{4}{*}{Dataset} & \multirow{4}{*}{Method} & \multicolumn{12}{c}{Time Span} \\
\cmidrule(lr){3-14} & & \multicolumn{4}{c}{Short} & \multicolumn{4}{c}{Medium} & \multicolumn{4}{c}{Long} \\
\cmidrule(lr){3-6} \cmidrule(lr){7-10} \cmidrule(lr){11-14} & & CC $\uparrow$ & R$^2$ $\uparrow$ & SE $\downarrow$ & AE $\downarrow$ & CC $\uparrow$ & R$^2$ $\uparrow$ & SE $\downarrow$ & AE $\downarrow$ & CC $\uparrow$ & R$^2$ $\uparrow$ & SE $\downarrow$ & AE $\downarrow$\\
\midrule
\multirow{8}{*}{Chewie} & Vanilla & 79.39  &  59.29 & 0.0572 & 49.32 & 61.38 & 36.14 & 0.1010 & 62.11 & 59.57 & 32.82 & 0.1143 & 64.95 \\
& KRSRL & 80.54 & 60.61 & 0.0537 & 49.32 & 63.32 & 37.51 & 0.0996 & 60.31 & 59.02 & 32.15 & 0.1157 & 66.33 \\
& MRNN & 85.83 & 71.21 & 0.0422 & 47.15 & 70.01 & 48.77 & 0.0910 & 54.86 & 63.26 & 36.45 & 0.1089 & 64.41 \\
& LFDA & 87.91 & 72.35 & 0.0354 & 49.54 & 74.58 & 55.12 & 0.0768 & 51.59 & 68.14 & 41.21 & 0.1028 & 58.58 \\
& CDNG & 88.02 & 75.59 & 0.0300 & 42.15 & 76.26 & 57.91 & 0.0693 & \textbf{42.15} & 68.91 & 43.35 & 0.0906 & 57.77 \\
& WDGRL & 87.35 & 69.79 & 0.0397 & 45.89 & \textbf{78.82} & \textbf{65.92} & \textbf{0.0489} & 47.14 & \textbf{80.09} & \textbf{60.03} & \textbf{0.0619} & \textbf{49.90} \\
& CND & 87.07 & 75.58 & 0.0348 & 47.94 & 77.78 & 56.54 & 0.0819 & 48.83 & 66.31 & 43.31 & 0.1060 & 56.05 \\
& \textbf{SSCDL} & \textbf{88.21} & \textbf{77.56} & \textbf{0.0297} & \textbf{39.50} & 77.53 & 59.66 & 0.0639 & 49.54 & 67.63 & 45.40 & 0.0880 & 56.73 \\
\midrule
\multirow{8}{*}{Mihili} & Vanilla & 68.07 & 45.84 & 30.05 & 53.07 & 67.66 & 49.17 & 33.05  & 54.07 & 48.49 & 22.48 & 54.36  & 61.33 \\
& KRSRL & 77.36 & 58.93 & 30.55 & 36.37 & 68.18 & 45.60 & 29.29 & 53.38 & 49.43 & 23.66 & 55.72 & 60.96 \\
& MRNN & 78.19 & 61.24 & 28.79 & 35.43 & 70.63 & 51.37 & 28.72 & 50.39 & 58.42 & 38.19 & 52.93 & 57.38 \\
& LFDA & 82.22 & 64.77 & 25.09 & 34.04 & 74.10 & 53.89 & 24.53 & 48.62 & 65.24 & 41.30 & 48.35 & 50.77 \\
& CDNG & 82.31 & 65.81 & 20.69 & 33.81 & 74.26 & 54.91 & 22.07 & 48.05 & 65.43 & 42.98 & 40.91 & 50.11 \\
& WDGRL & 82.52 & 65.49 & 27.06 & 35.82 & 71.28 & \textbf{55.58 }& 23.89 & 48.69 & 55.21 & 40.56 & 50.22 &  51.89\\
& CND & 81.67 & 65.76 & 21.64 & 34.40 & 74.24 & 54.60 & 22.80 & 48.45 & 64.88 & 42.78 & 42.48 & 50.67 \\
& \textbf{SSCDL} & \textbf{82.34} & \textbf{66.67} & \textbf{18.20} & \textbf{33.12} & \textbf{74.43} & 54.95 & \textbf{20.79} & \textbf{47.27} & \textbf{65.53} & \textbf{43.44} & \textbf{36.54} & \textbf{49.57} \\
\midrule
\multirow{8}{*}{Jango} & Vanilla & 77.91 & 57.53 & 14.11 & 68.00 & 74.37 & 52.45 & 14.29 & 66.76 & 71.93 & 48.64 & 15.88 & 72.89 \\
& KRSRL & 77.08 & 56.61 & 14.22 & 68.70 & 74.23 & 51.55 & 15.02 & 65.94 & 72.36 & 49.36 & 15.79 & 70.50 \\
& MRNN & 81.03 & 65.79 & 10.19 & 68.16 & 77.46 & 65.30 & 12.76 & 67.84 & 74.35 & 53.79 & 14.16 &  70.05\\
& LFDA & 84.39 & 70.66 & 8.576 & 69.90 & \textbf{82.28} & 65.30 & 9.877 & 70.55 & 80.86 & 58.43 & 13.48 & 72.57 \\
& CDNG & 84.71 & 71.63 & 7.117 & 67.91 & 81.51 & 66.01 & 8.870 & 65.76 & 65.43 & 42.98 & 11.42 & 69.57 \\
& WDGRL & 83.26 & 68.05 & \textbf{6.214} & \textbf{62.96} & 78.32 & 64.98 & \textbf{7.870} & \textbf{60.14} & 78.15 & 62.33 & 13.91 & 74.11 \\
& CND & 84.95 & 71.81 & 7.556 & 69.50 & 81.34 & 65.93 & 8.711 & 70.30 & 80.15 & 62.93 & 11.30 & 72.29 \\
& \textbf{SSCDL} & \textbf{85.04} & \textbf{72.26} & 6.798 & 65.70 & 81.72 & \textbf{67.58} & 8.526 & 62.96 & 80.48 & 63.57 & 9.772 & 66.03 \\
\bottomrule
\end{tabularx}
\end{table*}
\begin{figure*}[t]
\centering
\includegraphics[width=0.9\linewidth]{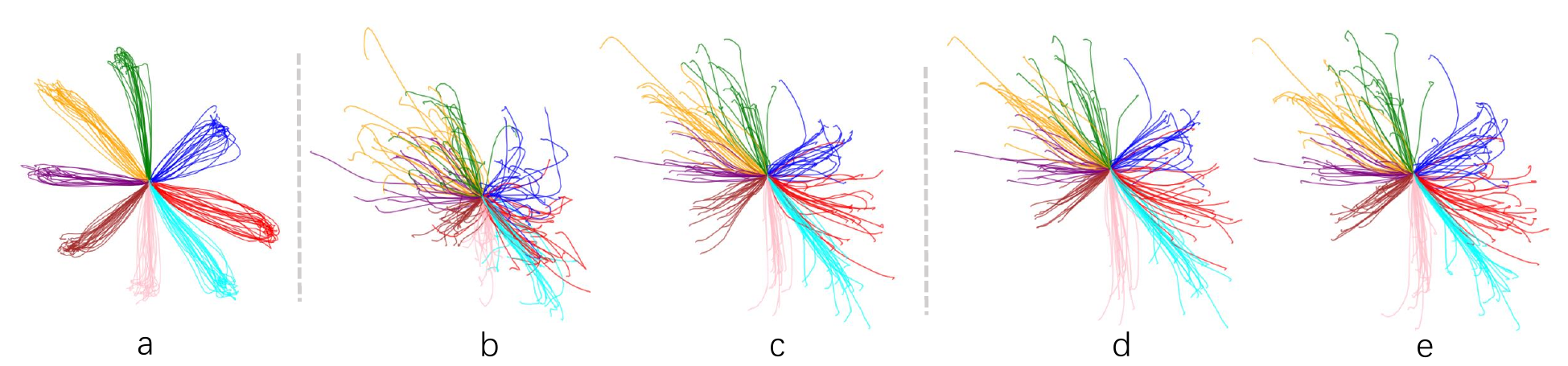}
\caption{Comparison of motor trajectories decoded by different models. Trajectories are color-coded by trial directions. (a) Ground truth; (b) Vanilla; (c) LFDA; (d) CND; (e) SSCDL.}
\label{fig:traj}
\end{figure*}

\subsubsection{Qualitative Results}
Trajectory visualizations in Fig.~\ref{fig:traj} further demonstrate the advantages of CND and SSCDL in decoding stability. The vanilla MLP baseline produces highly erratic trajectories, while LFDA partially alleviates this issue but still exhibits noticeable overshooting. The CND backbone already generates smooth and compact trajectories, demonstrating the effectiveness of the proposed decoder architecture. SSCDL further maintains similarly stable and accurate trajectories that closely match the ground-truth movements. Such stable and precise decoding is essential for reliable and intuitive BMI control.

\subsubsection{Fine-grained Analysis}
Beyond overall velocity prediction, SSCDL explicitly disentangles velocity into direction and speed components, enabling a more detailed evaluation of decoding performance. We therefore report the Speed Error (SE) and the Angle Error (AE), and these results validate the effectiveness of the CDG module in capturing neural population shifts related to specific kinematic components.

\subsection{Ablation Study (RQ2)}
To analyze the contribution of each module in SSCDL, we conduct ablation experiments on session C0--C1. The results are summarized in Table~\ref{tab:exp-ablation}. From these results, we draw three key observations:
\begin{enumerate}
\item \textbf{CND provides a strong backbone.} Directly applying a decoder trained on the source domain leads to a significant performance drop on the target domain, indicating severe neural drift. By introducing neural signal masking (NSM), teacher-student consistency constraint, and the EMA-based teacher update, CND substantially improves cross-day decoding performance. These improvements demonstrate that CND effectively learns more stable neural representations. Combined with the results in Table~\ref{tab:exp-main}, this confirms SSD as a robust backbone for cross-day neural decoding.
\item \textbf{CDG improves generalization through disentangled learning.} CDG enhances generalization by decomposing velocity into complementary components: direction, speed, and overall velocity. The velocity branch provides balanced predictions, while the direction and speed branches specialize in their respective motor parameters. Even when using the baseline backbone, CDG improves performance compared with the single-velocity decoder. When combined with CND, the improvements become more pronounced, indicating that CDG is compatible with robust representation learning.
\item \textbf{Both modules are essential and beneficial.} While CND and CDG individually improve cross-day generalization, their integration in SSCDL achieves the best overall performance. SSD contributes stable neural representations that mitigate neural drift, whereas CDG introduces parameter-specific specialization through disentangled learning. These two modules together provide a robust framework for long-term neural decoding, enabling more stable and accurate BMI control.
\end{enumerate}

\begin{table}[t]
\centering
\caption{Ablation results of SSCDL on session C0--C1}\label{tab:exp-ablation}
\label{tab:3}
\begin{tabular}{lcccc}
\toprule
Method & CC $\uparrow$ & $R^2$ $\uparrow$ & AE $\downarrow$ & SE $\downarrow$ \\
\midrule
Baseline & 79.39 & 59.29 & 49.32 & 0.0572 \\
Baseline + EMA & 80.01 & 60.53 & 48.37 & 0.0555 \\
Baseline + NSM & 84.35 & 71.79 & 48.24 & 0.0457 \\
Baseline + NSM + EMA & 86.35 & 73.79 & 48.04 & 0.0421 \\
SSD & 87.07 & 75.58 & 47.94 & 0.0348 \\
\midrule
Baseline (Velocity) & 79.39 & 59.29 & 49.32 & 0.0572 \\
Baseline (Direction) & N/A & N/A & 44.37 & 0.1040 \\
Baseline (Speed) & N/A & N/A & 80.54 & 0.0385 \\
CoDG & 84.20 & 68.52 & 43.95 & 0.0374 \\
\midrule
SSD (Velocity) & 87.07 & 75.58 & 47.94 & 0.0348 \\
SSD (Direction) & N/A & N/A & 39.69 & 0.1687 \\
SSD (Speed) & N/A & N/A & 98.81 & 0.0312 \\
SSCDL & \textbf{88.21} & \textbf{77.56} & \textbf{39.50} & \textbf{0.0297} \\
\bottomrule
\end{tabular}
\end{table}
\begin{figure}[t]
\centering
\includegraphics[width=0.95\linewidth]{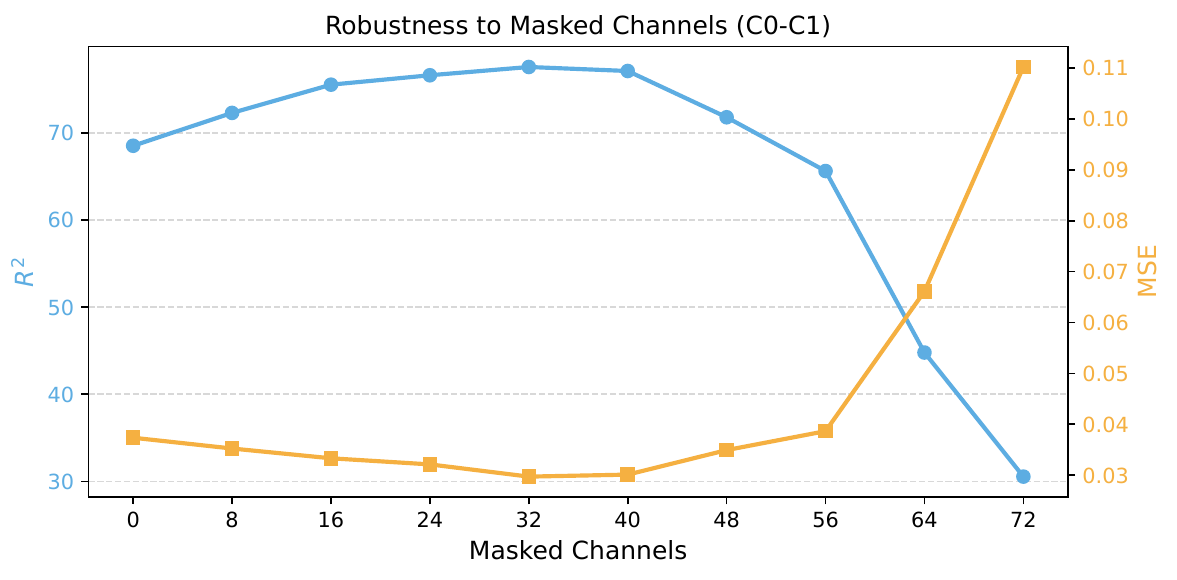}
\caption{Robustness Analysis on Number of Masked Channels.}
\label{fig:mask}
\end{figure}

\subsection{Robustness Analysis (RQ3)}
\subsubsection{Robustness Across Tasks, Subjects, and Timescales}
As shown in Table~\ref{tab:exp-main}, SSCDL demonstrates consistent generalization across different neural decoding tasks (CO, ISO, and RT), multiple subjects, and temporal spans of up to three months without any recalibration. These results indicate that the proposed framework maintains stable decoding performance under diverse experimental conditions, highlighting its scalability and suitability for long-term BMI applications.

\subsubsection{Robustness to Masked Channels}
To further evaluate robustness to neural signal corruption, we analyze the effect of channel masking on session C1 (Fig.~\ref{fig:mask}). The best performance is achieved when $32$ channels are masked, suggesting that moderate masking can improve representation learning by encouraging the model to rely on distributed neural information. When the number of masked channels increases to $48$, performance slightly decreases but remains stable. However, masking $64$ channels leads to a substantial degradation in performance, indicating the practical limit of tolerable channel occlusion.

\subsubsection{Robustness on Real-World Human Dataset}
We further evaluate the robustness of SSCDL on a real-world \textbf{human robotic effector dataset} from the FALCON benchmark~\cite{focal}. In this dataset, the source domain consists of recordings from 01-01, where the participant performs a virtual arm reaching task. The results in Table~\ref{tab:exp-human} demonstrate that SSCDL consistently outperforms existing methods across multiple target domains. These findings further validate the robustness and practical applicability of the proposed framework for real-world human–robot interaction scenarios.

\begin{table}[t]
\centering\caption{Robustness Analysis on the Human Dataset}
\label{tab:exp-human}
\begin{tabular}{lcccc}
\toprule
Target Domain & Vanilla & MRNN & LFDA & \textbf{SSCDL} \\
\midrule
01-08 & 10.43 & 11.42 & 12.35 & \textbf{13.54} \\
01-20 & 9.14  & 10.53 & 11.37 & \textbf{12.79} \\
02-09 & 8.76  & 9.56  & 11.10 & \textbf{12.03} \\
\bottomrule
\end{tabular}
\end{table}
\section{Discussion}
\subsubsection{Computational Cost} 
While SSCDL introduces a three-model ensemble, the computational overhead is moderate: inference for 1000 samples takes about 0.24s, with 31k parameters, comparable to or smaller than existing methods (Table~\ref{tab:cost}). These characteristics make SSCDL practically feasible for current BMI devices.

\begin{table}[t]
\centering
\caption{Computational Cost Analysis}
\label{tab:cost}
\begin{tabular}{lcccc}
\toprule
Metric & Static & MRNN & LFDA & Ours (Parallelized) \\
\midrule
Parameters & 11k & 42k & 28k & 31k \\
Inference Time & 0.17s & 0.41s & 0.27s & 0.24s \\
\bottomrule
\end{tabular}
\end{table}

\subsubsection{Limitations} 
Currently, SSCDL operates offline, limiting immediate application to real-time control. 
Future work will focus on enabling low-latency online decoding, reducing model complexity via pruning or distillation, and integrating multi-modal sensory feedback to further improve generalization across tasks, subjects, and sessions.

\subsubsection{BMIs for Human-Centered Robotics and Human-Robot Interaction} 
BMIs enable long-term, human-centered robotic control by translating neural activity into actionable commands for prostheses, exoskeletons, and teleoperation systems. By mitigating neural drift and parameter-specific variability, SSCDL enables stable and fine-grained neural decoding. The learned disentangled representations further support long-term cross-day and cross-subject control, facilitating adaptive and continuous human–robot interaction.

\section{Conclusion}
In this work, we present SSCDL, a neural decoding framework designed to improve cross-day robustness using only single-day training data. The proposed framework integrates CND with CDG to learn stable and transferable neural representations. CND leverages a self-supervised teacher–student paradigm to enforce prediction consistency under neural perturbations, while CDG decomposes velocity into complementary components (velocity, direction, and speed) to better capture parameter-specific neural dynamics. By mitigating neural drift and parameter-specific variability, SSCDL enables stable and fine-grained neural decoding while learning disentangled neural representations. This capability supports long-term cross-day and cross-subject BMI control, facilitating adaptive and continuous human–robot interaction in applications such as neurorehabilitation, human motor augmentation, and human-centered robotics.

\section{Acknowledgment}
This work was supported in part by the National Natural Science Foundation of China under Grant (U25D9015, 62336007), in part by the Starry Night Science Fund of the Zhejiang University Shanghai Institute for Advanced Study under Grant SN-ZJU-SIAS-002, in part by the Fundamental Research Funds for the Central Universities.
The authors acknowledge the use of Gemini (Google) for text polishing to improve the readability of this manuscript.

\bibliographystyle{IEEEtran}
\bibliography{refs}

\end{document}